\documentclass[sigconf]{acmart}
\AtBeginDocument{%
  }

\setcopyright{acmlicensed}
\copyrightyear{2018}
\acmYear{2018}
\acmDOI{XXXXXXX.XXXXXXX}
\acmConference[]{}{}{}
\acmISBN{978-1-4503-XXXX-X/2018/06}

\usepackage{xcolor}
\usepackage{listings}
\begin{document}

\title{Beyond Play \& Pause: Untwist – Turning GPT-4o’s Spatial Weakness into a Strength for In-Depth Interactive Video Learning}

\author{Sajad Goudarzi}
\email{sgoudar@clemson.edu}
\orcid{0009-0000-0682-1828}
\affiliation{%
  \institution{Clemson University}
  \city{Clemson}
  \state{SC}
  \country{USA}
}

\author{Samaneh Zamanifard}
\affiliation{%
  \institution{Clemson University}
  \city{Clemson}
  \country{USA}}
\email{szamani@clemson.edu}

\renewcommand{\shortauthors}{Goudarzi and Zamanifard}

\begin{abstract}
Traditional video-based learning remains passive, offering limited opportunities for users to engage dynamically with content. While current AI-powered tools offer transcription and summarization, they lack real-time, region-specific interaction capabilities. This paper introduces Untwist, an AI-driven system that enables interactive video learning by allowing users to ask questions about the entire video or specific regions using a bounding box, receiving context-aware, multimodal responses. By integrating GPT APIs with Computer Vision techniques, Untwist extracts, processes, and structures video content to enhance comprehension. Our approach addresses GPT-4o’s spatial weakness by leveraging annotated frames instead of raw coordinate data, significantly improving accuracy in localizing and interpreting video content. This paper describes the system architecture, including video pre-processing and real-time interaction, and outlines how Untwist can transform passive video consumption into an interactive, AI-driven learning experience with the potential to enhance engagement and comprehension.
\end{abstract}



\received{20 February 2007}
\received[revised]{12 March 2009}
\received[accepted]{5 June 2009}

\maketitle

\section{Introduction}
The integration of artificial intelligence (AI) into education has significantly transformed learning experiences, making them more accessible and personalized \cite{zamanifard2025undergraduate,pelaez2024impact,abd2023large}. Among various AI-driven advancements, educational videos have become a powerful medium for conveying complex concepts through rich visual and auditory stimuli \cite{moss1983video}. Videos enhance comprehension, engagement, and retention by providing dynamic content that caters to different learning styles \cite{berk2009multimedia,kosterelioglu2016student}. As a result, online learning platforms and digital classrooms have increasingly adopted video-based instruction to supplement traditional teaching methods \cite{paramesti2021effect}. However, despite their benefits, most educational videos remain passive, offering limited opportunities for learners to interact with the content in real-time. Students watching instructional videos often lack immediate clarification of doubts, making it difficult to fully grasp intricate concepts without external resources or instructor guidance \cite{mitrovic2016reflective,dimitrova2017using}.

LLMs, such as ChatGPT, have recently emerged as transformative tools in education, offering personalized tutoring, adaptive assessments, and automated content generation \cite{pan2024integrating,mannekote2024large}. Their ability to process and generate natural language enables them to assist students with explanations, answer questions, and enhance engagement in self-directed learning environments \cite{fagbohun2024beyond,biri2023assessing}. While LLMs have demonstrated their efficacy in text-based interactions, their application in video-based learning remains limited. Current AI-powered educational tools either focus on passive video transcription, captioning, or summarization \cite{zhang2023video} but lack the capability for real-time, interactive engagement with video content. 

Recent advances in multimodal AI have enabled LLMs to process texts, images, and videos for improved semantic understanding and reasoning \cite{zhou2404survey,ko2023large}. In video question answering (VideoQA), LLMs analyze frames and generate responses based on extracted information \cite{zhang2023video}. However, these models, lack real-time, user-driven interactions, leaving users as passive consumers rather than active participants in video-based learning.

In this work, we introduce Untwist,  an interactive AI-powered system that dynamically engages users with video content. Our system enables learners to select specific video segments and pose questions, with an LLM providing context-aware responses based on the selected portion. Unlike prior approaches that focus solely on static text-based queries or pre-defined video captions, our method bridges the gap between video comprehension and interactive learning by leveraging multimodal AI to support real-time, personalized engagement. Our key contributions are as follows:
\begin{itemize}
    \item We introduce a novel interactive AI-powered system, Untwist, that allows users to ask questions about the entire video or specific regions using a bounding box, receiving context-aware, multimodal responses.
    \item We propose an adaptive video comprehension framework that processes user-selected video segments and generates responses using multimodal AI technique, improving the depth of video-based learning interactions.
    \item We introduce a novel approach to handling GPT-4o’s spatial limitations by replacing raw coordinate-based input with annotated frames, significantly improving the model’s ability to localize and interpret video content accurately.
    \item We demonstrate the feasibility of real-time video-LLM interaction and outline potential applications in education, training, and beyond.
\end{itemize}
\section{Related Work}
\subsection{Multimedia Learning in Education}
The integration of video into educational settings has transformed the way knowledge is imparted and absorbed. Over the past few decades, video has become an essential medium for education due to its ability to convey complex information in an engaging and accessible format \cite{moss1983video}. However, traditional video-based learning remains largely passive, often requiring additional instructional tools or assessments to reinforce engagement and ensure comprehension. 

One of the key advantages of video in education is its ability to engage multiple senses, helping students absorb content more effectively than traditional textual or auditory materials. Videos provide both visual and auditory stimuli, making them particularly effective for presenting complex concepts that might be difficult to grasp through text alone \cite{berk2009multimedia,kosterelioglu2016student}. For instance, in fields such as medicine, engineering, and science, videos can demonstrate procedures, experiments, or simulations that would be impossible to replicate in a classroom setting. This allows students to experience and interact with content in a way that static text cannot \cite{mayer2008applying}. Educational videos also cater to a wide range of learning styles. Visual learners benefit from seeing concepts in action, while auditory learners gain insights from spoken explanations. Videos can also provide real-life examples that appeal to kinesthetic learners, who learn best through movement and action. By addressing these diverse learning preferences, video-based instruction makes learning more inclusive and accessible \cite{moreno2007interactive,lackmann2021influence}. Additionally, videos offer flexibility, allowing students to learn at their own pace by pausing, replaying, and reviewing content. This enhances satisfaction and improves learning outcomes, particularly in self-paced educational environments \cite{lo2019impact,mahmud2011dissection}.

Despite these advantages, there are notable limitations to video-based education. One significant challenge is the potential for students to feel disconnected in asynchronous learning environments, as videos lack personal interaction and real-time feedback, which can hinder engagement and comprehension \cite{teoh2023machine}. While some platforms mitigate this limitation by incorporating quizzes or discussion forums, they do not offer real-time, context-aware responses tailored to a learner’s specific focus within a video. Furthermore, the effectiveness of video learning can vary depending on content presentation methods (e.g., narrated versus subtitled formats) \cite{tarchi2021learning}. These limitations highlight the need for AI-driven solutions that enable interactive, contextually aware engagement during video learning.

\subsection{Large Language Models (LLMs) in Education}
As video-based learning continues to expand, its limitations, such as the lack of real-time, context-aware feedback and personal interaction, highlight the need for more interactive solutions \cite{dimitrova2017using}. The integration of AI-driven tools such as LLMs offers new opportunities to enhance interactivity and comprehension. LLMs, such as ChatGPT, represent a significant advancement in artificial intelligence, particularly in natural language processing \cite{pan2024integrating,mannekote2024large}. In education, they provide personalized learning experiences, adaptive assessments, and enhanced instructional support, addressing diverse learning needs \cite{pan2024integrating,mannekote2024large}. Their ability to analyze and generate text makes them highly valuable for various educational applications, including question generation, automated grading, and providing rich explanations beyond traditional materials. This adaptability has been shown to improve student confidence, foster deeper engagement, and support self-directed learning across multiple disciplines \cite{fagbohun2024beyond,biri2023assessing,zhui2024impact}.

LLMs have been particularly recognized for their role as supplementary educational resources, helping students access explanations that go beyond conventional textbooks. For example, in medical education, students leverage LLMs to explore complex topics with greater clarity, increasing their confidence in discussing intricate subjects \cite{biri2023assessing}. Additionally, a nationwide survey among medical educators highlights the potential of LLMs in creating dynamic learning environments, emphasizing the need for educator training programs to effectively integrate these technologies into teaching \cite{roy2024assessing}. Beyond structured disciplines, LLMs play a crucial role in language education, particularly in enhancing personalized learning experiences. They support critical language skills such as speaking and writing, enabling learners to receive tailored feedback and refine their linguistic abilities \cite{baskara2023navigating,angel2023performance}.

Building on their success in text-based learning, LLMs are now being explored for AI-driven video understanding. Recent advancements show their application in semantic video segmentation, where they improve object recognition and classification by capturing subtle semantic nuances \cite{zhou2404survey}. LLMs also facilitate multimodal learning by processing visual, auditory, and textual data, allowing for deeper video comprehension than traditional methods \cite{zhang2023video}. Furthermore, LLMs enhance video question answering (VideoQA) by leveraging temporal and causal reasoning, enabling them to understand relationships between video content and user queries \cite{ko2023large}. However, while these approaches enhance automated video comprehension, they do not yet enable interactive, user-driven engagement with video content, a gap that our work seeks to address.

Beyond video comprehension, LLMs are being utilized to generate dynamic content, such as adaptive dialogues in video games and automated script creation for educational videos \cite{akoury2023framework}. Despite these advancements, existing LLM-powered educational tools primarily focus on text-based interactions and passive video analysis, lacking the ability to provide real-time, user-driven engagement with multimedia content. Unlike prior work that focuses on either text-based question answering or static video captioning, our approach introduces an interactive AI system that enables users to directly engage with video content by selecting specific segments and receiving contextualized LLM-generated responses. In this work, we address this gap by enabling direct interaction with video, allowing users to engage in AI-powered conversations tailored to their selected content.

\section{System Design and Architecture}
Building on the opportunities and limitations of existing LLM applications in video-based learning, this section details the design of  Untwist, a system developed to bridge the gap by providing real-time, user-driven interaction. Untwist is an AI-powered interactive learning system designed to transform passive video consumption into an engaging, interactive learning experience. The system allows users to ask questions about video content in two ways: by asking general questions about the entire video, or by asking detailed questions about specific regions by using a bounding box.

By integrating LLMs with video processing techniques, Untwist enables real-time, AI-driven explanations tailored to user-selected content. The system consists of multiple interconnected components that extract, process, and generate responses based on video data, user interactions, and AI-generated insights.

\begin{figure*}[h]
    \centering
    \includegraphics[width=0.6\textwidth]{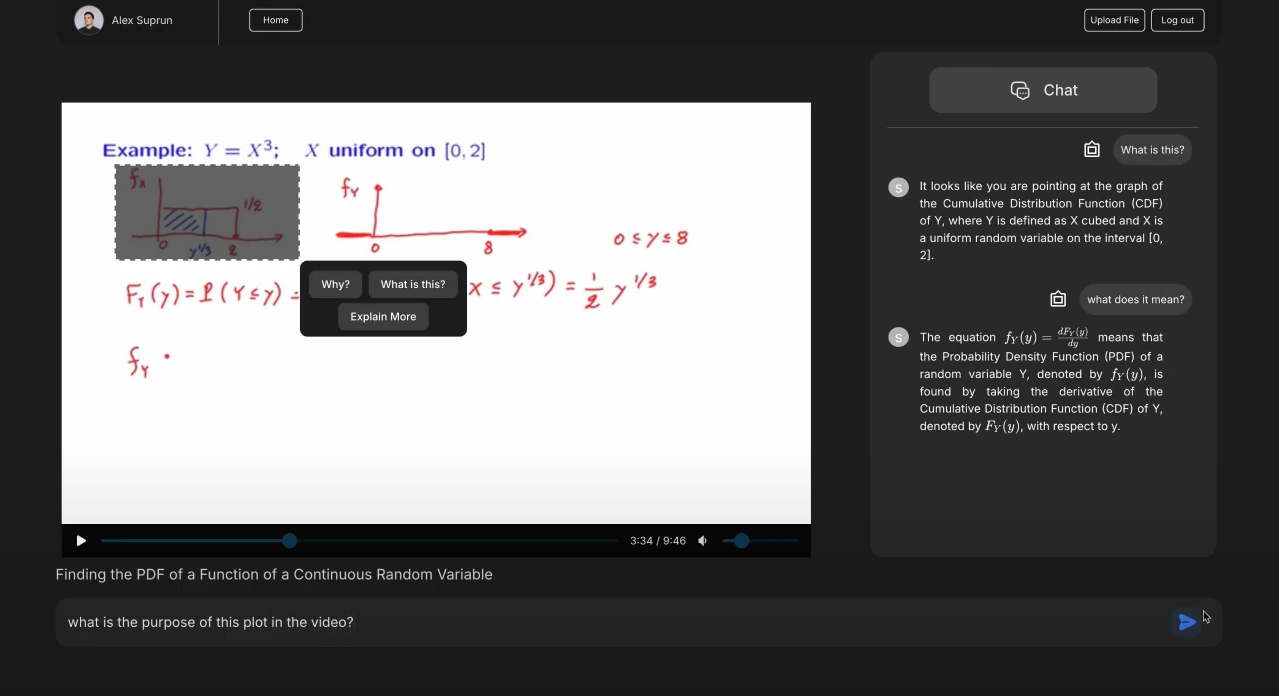} 
    \caption{The User Interface of Untwist, enabling interactive video-based learning through AI-driven explanations.}
    \label{fig:system_ui}
\end{figure*}

Figure~\ref{fig:system_ui} illustrates the user interface of Untwist, where users can interact with video content dynamically. The right panel features a chatbot that provides AI-generated responses based on the user’s selection or general inquiries. Users can highlight parts of the video, triggering the system to process the selection and generate relevant explanations.

\begin{figure*}[h]
    \centering
    \includegraphics[width=\linewidth]{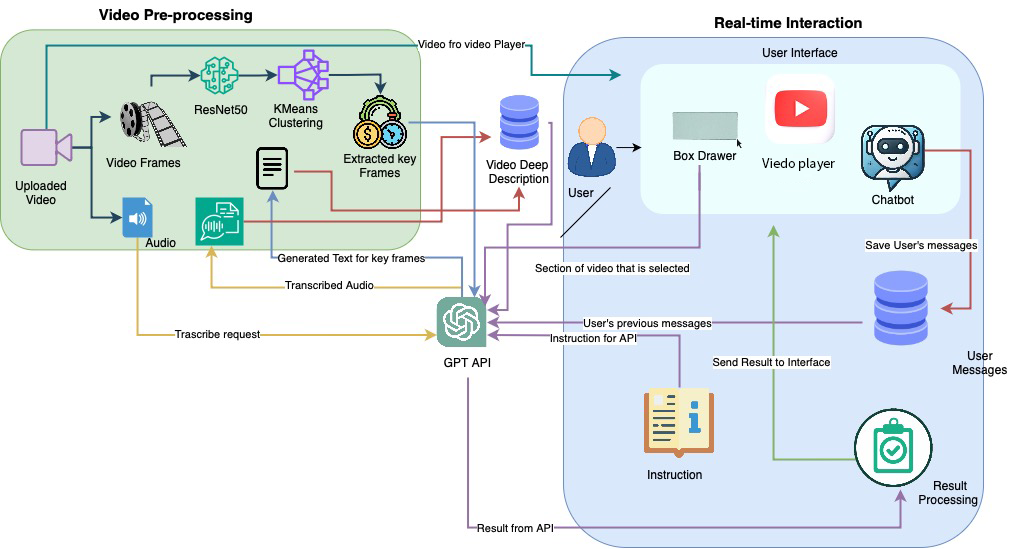}  
    \caption{System Architecture of Untwist.}
    \label{fig:system_architecture}
\end{figure*}

This section describes the architecture of Untwist, detailing its key components, workflow, and integration of LLMs for real-time video interaction. The architecture of Untwist is shown in Figure \ref{fig:system_architecture}, it consists of two main components: Video Pre-processing and Real-Time Interaction, each playing a crucial role in processing video content, facilitating user interactions, and generating meaningful AI-driven responses.

\subsection{Video Pre-processing}
The video pre-processing component is responsible for analyzing video content before user interaction. This module extracts meaningful text, speech, and key visual elements from videos to enhance real-time engagement. The key submodules include:
\begin{itemize}
    \item Transcription Module: The system extracts spoken content from the video and utilizes OpenAI APIs to transcribe speech into text, enabling further processing and integration with video analysis.
    \item Key Frame Extraction \& Optimization: Identifies and extracts important frames that contain diagrams, equations, or key moments. It uses computer vision (CV) techniques to implement an adaptive frame reduction strategy to minimize the number of extracted frames while ensuring key content is preserved. Optimizes the number of frames N based on content complexity, avoiding redundancy while maintaining accuracy. Ensures a balance between computational efficiency and information retention, improving system responsiveness and scalability.
    \item Video Deep Description: The system leverages the GPT-4o model via the OpenAI API to extract and interpret key content from video frames, generating a structured and coherent summary of essential elements. It integrates multiple data sources, including transcribed audio, keyframe extractions, and frame content analysis, to construct a context-aware narrative that logically connects video concepts. This approach ensures that key ideas, equations, and explanations are systematically linked, enabling the AI to generate accurate and contextually relevant responses to user queries. Additionally, LLMs refine and structure the generated summary before it is sent to the chatbot, enhancing clarity and coherence for a more seamless user interaction.
\end{itemize}

\subsection{Real-time Interaction}
The real-time interaction component allows users to engage dynamically with the video by selecting segments and receiving AI-generated responses. The core submodules include:
\begin{itemize}
    \item Video Player \& Box Drawer: Displays the video and allows users to interact with that in ordinary manner. It also gives the user the ability to highlight specific parts (e.g., equations, graphs, or subtitles). The selection of the bounding box is captured with timestamps and coordinates for AI processing.
    \item Chatbot Interaction: The system enables users to engage with the chatbot by asking general questions about the video or posing specific inquiries by selecting a particular segment using a bounding box. It integrates the user's previous interactions, the current video deep description, and general topic knowledge to generate context-aware, relevant responses. This approach ensures a more personalized and insightful interaction, enhancing the user's learning experience.

    \item User Query History Storage: The system maintains a record of user queries and AI responses, leveraging this history to generate personalized replies tailored to the user's knowledge and past interactions. This ensures continuity in conversations and enhances the relevance of responses.
\end{itemize}

This module enables real-time processing of user selections, seamlessly mapping video content to AI-generated responses for accurate and context-aware answers.

\section{Implementation Details}
In this section, we detail the technical implementation of Untwist, covering the video deep description generator, the interactive video player, and the backend context handling and response generation.

\subsection{Video Deep Description Generator}
The deep description generator is implemented in Python and leverages both deep learning and classical machine learning libraries to extract and summarize key video content. Our pipeline consists of the following technical modules:

\begin{enumerate}
    \item \textbf{Frame Extraction and Pre-processing:}
    \begin{itemize}
        \item Frames are extracted at a fixed interval of one frame every two seconds using the Python \texttt{OpenCV} library, which is appropriate given the relatively static nature of educational videos.
        \item Each extracted frame is resized to 224$\times$224 pixels and normalized using the standard ImageNet mean and standard deviation values.
    \end{itemize}

    \item \textbf{Feature Extraction:}
    \begin{itemize}
        \item Preprocessed frames are forwarded through a pretrained ResNet50 (available via \texttt{torchvision.models}) with its final fully connected layer removed. This produces 2048-dimensional feature vectors for each frame.
    \end{itemize}
    
    \item \textbf{Key Frame Selection:}
    \begin{itemize}
    \item Feature vectors are clustered using the KMeans algorithm from \texttt{scikit-learn}. The number of clusters ($K$) is determined dynamically using an elbow method based on video duration and content variance.
    \item For each cluster, the frame whose feature vector is closest to the computed centroid is selected as the representative key frame.

    \end{itemize}
    
    \item \textbf{Audio Transcription:}
    \begin{itemize}
    \item The audio track is extracted from the video using the \texttt{MoviePy} library's \texttt{VideoFileClip} function and then transcribed using the OpenAI Whisper API.
    \end{itemize}
    
\item \textbf{LLM Integration:}
\begin{itemize}
    \item For each representative key frame, a structured prompt is created by combining the transcribed audio with the corresponding image frame. The prompt instructs GPT-4o to analyze the visual content and return a JSON object describing the frame. An example prompt is:

\begin{lstlisting}
"This is the transcribed audio from an educational video: 
[transcribed_text]"

"This is an image frame from the educational video, captured at 
[timestamp] seconds: [image_file]"

"Please analyze the image and describe its content. 
Return your answer in the following JSON format:
{
  'math': 'math expression in LaTeX',
  'text': 'descriptive text present in the image',
  'graph': 'description of any graph observed',
  'other_shapes': 'description of any other shapes or figures',
  'additional_info': 'any additional observations'
}"
\end{lstlisting}
    \item The prompt is sent to GPT-4o, which returns a JSON object detailing the content of the image. This output is then integrated into the overall video description for further context-aware processing.
\end{itemize}
\end{enumerate}

\subsection{Interactive Video Player}
The front-end is built using the Next.js framework, integrating modern React components and real-time communication protocols. Key technical details include:

\begin{enumerate}
    \item \textbf{UI and Annotation:} 
    \begin{itemize}
        \item The video player is implemented with the HTML5 \texttt{<video>} element, augmented by custom React components.
        \item The \texttt{react-konva} library is used to render an interactive canvas overlay, enabling users to draw bounding boxes over regions of interest.
    \end{itemize}
    
\item \textbf{Event Handling and Data Capture:}
\begin{itemize}
    \item Mouse events are tracked to capture the bounding box coordinates relative to the video frame.
    \item Box coordinates are computed relative to the video player's display size, ensuring consistency across different player dimensions.
    \item The current timestamp is extracted directly from the video player API.
    \item The bounding box coordinates, the corresponding timestamp, and the user message are packaged into JSON objects and transmitted to the backend.

\end{itemize}

    \item \textbf{Real-Time Communication:}
    \begin{itemize}
        \item A WebSocket connection, established via the \texttt{socket.io} library, enables low-latency, bidirectional communication between the client and the server.
    \end{itemize}
\end{enumerate}

\subsection{Context Handling and Response Generation}
The backend, implemented in Python using the Django framework, orchestrates multimodal data aggregation and communication with the GPT API. Its core functions are:

\begin{enumerate}
    \item \textbf{Data Aggregation and Pre-processing:}
    \begin{itemize}
        \item Upon receiving a query, the backend retrieves the corresponding video frame using the provided timestamp.
        \item The \texttt{OpenCV} library is used to overlay the received bounding box on the frame, generating an annotated image.
    \item The annotated frame, precomputed deep description, and the user's interaction history (stored in MongoDB, a NoSQL database) are compiled into a JSON payload.

    \end{itemize}
    
\item \textbf{LLM Query:}
\begin{itemize}
    \item The compiled payload is sent to GPT-4o via continuous chat completion provided by OpenAI Python library.
\end{itemize}

\item \textbf{Response Delivery:}
\begin{itemize}
    \item Responses are delivered back to the client over a WebSocket connection.
    \item All interactions are logged to maintain session continuity and enable further context refinement.
\end{itemize}
\end{enumerate}

\subsection{Deployment:} The complete system is deployed on a cloud platform running the Ubuntu operating system. The architecture is fully containerized using Docker, with separate containers for the frontend, the backend, an Nginx web server, and MongoDB. A CI/CD pipeline via GitHub Actions facilitates continuous integration, testing, and deployment.

Overall, these technical implementations ensure that Untwist can process video content in real-time, provide detailed and context-aware descriptions, and support interactive querying with low latency.

\section{Design Rationale: Annotated Frames vs. Raw Coordinate Data}

A critical design decision in Untwist was determining the most effective way to convey the user-selected region of interest to the GPT-4o model. Initially, we experimented with sending the entire video frame along with raw numerical bounding box coordinates (e.g., \texttt{"Coordinates: x=100, y=150, width=200, height=100"}) as part of the input. However, our empirical study, conducted on a dataset of 200 synthetic images, revealed that GPT-4o struggles to effectively interpret and utilize raw coordinate data. We also conducted the study using the O1 model, which has enhanced reasoning capabilities. However, it required significantly longer processing times, approximately 10 times slower than GPT-4o, making it unsuitable for educational applications where users expect real-time responses. Additionally, the cost of running the study or using the O1 model in practice was substantially higher than GPT-4o. Despite these differences, the results were nearly identical to GPT-4o model for a dataset of 20 images, indicating that the increased processing time and cost did not provide a significant advantage in this context.

To isolate the effect of spatial representation, the synthetic images were generated with randomized dimensions. Specifically, image widths and heights were randomly selected between 1200 and 2000 pixels. Each image features a white background with black text to maximize text visibility and reduce potential letter recognition issues within the annotated region. This controlled setup was designed so that the primary variable under investigation was GPT-4o’s spatial understanding when provided with numerical coordinates versus a drawn bounding box.

We then created two separate prompting strategies:
\begin{enumerate}
    \item In the first approach, we sent the annotated image (i.e., the full image with the red bounding box drawn directly on it) along with a prompt asking GPT-4o to extract and return the text contained within the red box.
    \item In the second approach, we provided the full image along with its height, width, and the raw numerical coordinates of the desired area. The prompt instructed GPT-4o to extract the text contained within the specified region.

\end{enumerate}
To measure the effectiveness of each approach, we developed a Python analysis script. For each test image, the script tokenizes both the GPT-4o output and the ground truth text (the text inside the box) by splitting on whitespace and converting to lowercase. It then computes:
\begin{itemize}
    \item \textbf{Precision:} The ratio of the number of common tokens (the intersection between the GPT-4o's output and the ground truth) to the total number of tokens in GPT-4o's response.
    \item \textbf{Recall:} The ratio of the number of common tokens to the total number of tokens in the ground truth.
    \item \textbf{F1 Score:} The harmonic mean of precision and recall.
\end{itemize}

Our results demonstrated a significant difference between the two methods. As shown in Table 1, the \textbf{Raw Coordinate (no box) Approach} resulted in substantially lower precision, recall, and F1 scores, whereas the \textbf{Annotated Frame (box) Approach} achieved significantly higher accuracy across all metrics.

These findings indicate that when GPT-4o is provided with a full image annotated with the drawn bounding box, it can reliably localize the region of interest and accurately extract the intended text. Conversely, providing raw coordinate data results in a significant drop in performance, as the model struggles to map numerical values to the correct image region.

\begin{figure}[ht]
    \centering
    \includegraphics[width=0.6\linewidth]{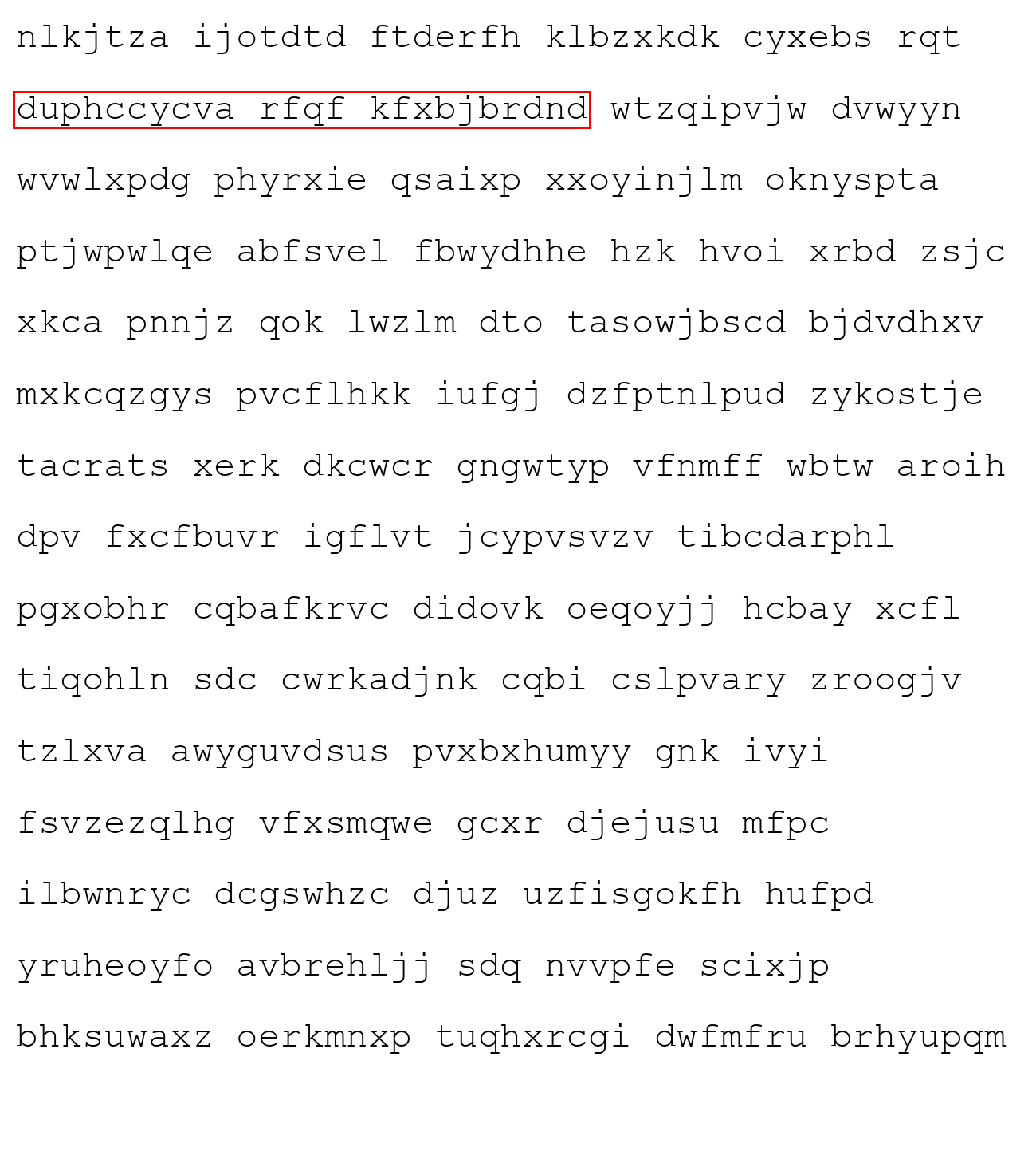}
    \caption{Example of a synthetic image with a randomly generated word and a red bounding box drawn over the region of interest.}
    \label{fig:annotated_example}
\end{figure}

Figure~\ref{fig:annotated_example} illustrates an example of an annotated image with a red bounding box, and Table~\ref{tab:Performance Comparison} summarizes the performance comparison between the two approaches.

\begin{table}
    \centering
    \begin{tabular}{lcc}
        \hline
        \textbf{Approach} & Raw Coordinate & Annotated Frame\\
        \hline
        \textbf{Precision (\%)}&5.19&84.82\\
         \textbf{Recall (\%)}&11.15&85.05\\
         \textbf{F1 Score (\%)}&6.54&84.92\\
        \hline
    \end{tabular}
    \caption{Comparison of Performance Between Annotated Frame and Raw Coordinate Approaches}
    \label{tab:Performance Comparison}
\end{table}

Based on these findings, our system adopts the annotated frame approach. By providing GPT-4o with a full image in which the region of interest is clearly marked by a drawn bounding box, we ensure that the model receives an unambiguous visual cue. This design choice significantly enhances GPT-4o's ability to generate accurate and contextually relevant responses, ultimately improving the interactive video learning experience.

\section{Applications and Use Cases}
Untwist's interactive video exploration framework opens up a wide range of real-world applications where AI-driven, context-aware feedback adds significant value. Below, we outline some key use cases across various domains.

\subsection{Education and E-Learning}
\paragraph{Flipped Classrooms.} Instructors can provide pre-recorded lectures for students to watch at their own pace. Using Untwist, students can highlight confusing diagrams or equations and receive real-time clarifications, reducing the burden on teachers to answer repetitive questions in class.

\paragraph{Supplemental Tutoring.} Students learning complex subjects, such as STEM disciplines, often require detailed explanations and examples. Untwist offers immediate, context-sensitive feedback to deepen understanding, complementing traditional textbooks and online discussion forums.

\paragraph{MOOC Platforms.} Massive open online courses can integrate Untwist to provide on-demand support at scale. Learners worldwide can interact with video content directly, pose questions, and resolve misunderstandings without waiting for dedicated instructor feedback.

\subsection{Corporate Training and Employee Onboarding}
\paragraph{Interactive Tutorials.} In corporate and industrial settings, training videos often cover intricate procedures or safety protocols. Untwist allows trainees to spotlight specific machine parts or workflow steps and instantly receive AI-driven clarifications, improving retention and reducing training time.

\paragraph{Compliance and Policy Videos.} Employees can select relevant parts of policy or regulatory videos and receive immediate guidance or clarifications on compliance procedures, ensuring consistent interpretations across the organization.

Together, these applications underscore the versatility of Untwist’s multimodal framework in enhancing video-based content across educational, professional, and entertainment domains.

\section{Challenges and Limitations}
While Untwist offers a promising avenue for enhancing video-based learning experiences through real-time interactions, several challenges and limitations remain:

\paragraph{Model Reliability.} Although LLMs such as GPT-4o exhibit strong linguistic and multimodal capabilities, they may still produce factually incorrect or hallucinated content when presented with ambiguous or highly specialized queries. This potential inaccuracy necessitates human oversight, particularly in critical educational or professional contexts.

\paragraph{Spatial Understanding.} As shown in our empirical evaluation, LLMs struggle to interpret raw numerical coordinates effectively. While providing annotated images with bounding boxes significantly improves performance, this approach increases the volume of data transferred to the model, potentially introducing latency or bandwidth constraints. Although specialized vision-language architectures handle bounding boxes natively, they often focus on detection or short-form responses rather than the open-ended, conversational reasoning provided by GPT-like models. Future multi-modal approaches that combine fine-grained spatial grounding with generative language capabilities may ultimately reduce reliance on fully annotated frames while preserving high-quality, in-depth explanations.

\paragraph{Frame Annotation.} While annotating the frame with a bounding box can significantly improve model performance, it does introduce potential sources of confusion. In educational videos, for instance, there may be limited color choices for the box, raising the risk that the bounding box color blends with on-screen elements of a similar hue. This overlap could decrease the model’s ability to accurately detect the annotated region. Careful color selection and design considerations are therefore essential to ensure the model can consistently recognize the true bounding box.

\paragraph{Computational Overheads.} Real-time video processing and LLM interactions are computationally intensive, especially for high-resolution or long videos. Scalability challenges arise with multiple users. Optimizing resource management, caching, and parallelization is essential for maintaining responsiveness.

\paragraph{Context Window Constraints.} Current large language models have a limited context window, which may prevent them from retaining long-ranging information about extended video content or many user queries within the same session. Though advanced LLMs provide larger context windows, there are still practical limits. Systems must therefore manage and summarize video content effectively, potentially leading to information loss or reduced context awareness.

\subsection{System Evaluation and Limitations}
The empirical evaluation presented in this paper is focused on the technical performance of the system, particularly the accuracy of GPT-4o’s spatial understanding when using annotated frames versus raw coordinate data. While our findings demonstrate the feasibility and technical efficacy of Untwist, it is important to acknowledge a key limitation: no formal user studies have been conducted to evaluate its real-world impact on learning outcomes. Therefore, claims regarding enhanced user engagement, comprehension, and retention are based on the system's design and are an area for future work.

\section{Conclusion}
This paper introduced Untwist, an AI-driven interactive video learning system that transforms passive video consumption into an active, engaging learning experience. We successfully addressed the primary weakness of traditional video-based learning, the lack of real-time, context-aware interaction, by integrating Large Language Models (LLMs) with computer vision techniques. A key technical contribution of this work is our novel approach to handling GPT-4o's spatial limitations. By providing the model with annotated frames that include a bounding box, we significantly improved its ability to accurately localize and interpret video content, achieving a much higher F1 score compared to using raw coordinate data. This finding is critical for building reliable, interactive systems for in-depth video learning.

Our empirical evaluation demonstrated the technical feasibility of real-time video-LLM interaction and outlined potential applications in education, corporate training, and beyond. However, the system faces challenges, including the high computational cost of real-time processing and the potential for factual inaccuracies in AI-generated responses. The limited context window of current LLMs also remains a constraint, as it may prevent the retention of long-ranging information over an extended video session.

Future work will focus on optimizing the system’s efficiency to handle longer and more complex videos. We will also conduct formal user studies to measure Untwist's real-world impact on learning outcomes, including student engagement, comprehension, and knowledge retention. By continuing to refine the system’s interactive and spatial reasoning capabilities, this work lays the foundation for more personalized and effective AI-assisted learning environments.

\bibliographystyle{ACM-Reference-Format}
\bibliography{sample-base}


\begin{thebibliography}{28}


\ifx \showCODEN    \undefined \def \showCODEN     #1{\unskip}     \fi
\ifx \showISBNx    \undefined \def \showISBNx     #1{\unskip}     \fi
\ifx \showISBNxiii \undefined \def \showISBNxiii  #1{\unskip}     \fi
\ifx \showISSN     \undefined \def \showISSN      #1{\unskip}     \fi
\ifx \showLCCN     \undefined \def \showLCCN      #1{\unskip}     \fi
\ifx \shownote     \undefined \def \shownote      #1{#1}          \fi
\ifx \showarticletitle \undefined \def \showarticletitle #1{#1}   \fi
\ifx \showURL      \undefined \def \showURL       {\relax}        \fi
\providecommand\bibfield[2]{#2}
\providecommand\bibinfo[2]{#2}
\providecommand\natexlab[1]{#1}
\providecommand\showeprint[2][]{arXiv:#2}

\bibitem[Abd-Alrazaq et~al\mbox{.}(2023)]%
        {abd2023large}
\bibfield{author}{\bibinfo{person}{Alaa Abd-Alrazaq}, \bibinfo{person}{Rawan AlSaad}, \bibinfo{person}{Dari Alhuwail}, \bibinfo{person}{Arfan Ahmed}, \bibinfo{person}{Padraig~Mark Healy}, \bibinfo{person}{Syed Latifi}, \bibinfo{person}{Sarah Aziz}, \bibinfo{person}{Rafat Damseh}, \bibinfo{person}{Sadam~Alabed Alrazak}, \bibinfo{person}{Javaid Sheikh}, {et~al\mbox{.}}} \bibinfo{year}{2023}\natexlab{}.
\newblock \showarticletitle{Large language models in medical education: opportunities, challenges, and future directions}.
\newblock \bibinfo{journal}{\emph{JMIR Medical Education}} \bibinfo{volume}{9}, \bibinfo{number}{1} (\bibinfo{year}{2023}), \bibinfo{pages}{e48291}.
\newblock


\bibitem[Akoury et~al\mbox{.}(2023)]%
        {akoury2023framework}
\bibfield{author}{\bibinfo{person}{Nader Akoury}, \bibinfo{person}{Qian Yang}, {and} \bibinfo{person}{Mohit Iyyer}.} \bibinfo{year}{2023}\natexlab{}.
\newblock \showarticletitle{A framework for exploring player perceptions of llm-generated dialogue in commercial video games}. In \bibinfo{booktitle}{\emph{Findings of the Association for Computational Linguistics: EMNLP 2023}}. \bibinfo{pages}{2295--2311}.
\newblock


\bibitem[Angel et~al\mbox{.}(2023)]%
        {angel2023performance}
\bibfield{author}{\bibinfo{person}{Mirana Angel}, \bibinfo{person}{Haiyi Xing}, \bibinfo{person}{Anuj Patel}, \bibinfo{person}{Amal Alachkar}, {and} \bibinfo{person}{Pierre Baldi}.} \bibinfo{year}{2023}\natexlab{}.
\newblock \showarticletitle{Performance of Large Language Models on Pharmacy Exam: A Comparative Assessment Using the NAPLEX}.
\newblock \bibinfo{journal}{\emph{bioRxiv}} (\bibinfo{year}{2023}), \bibinfo{pages}{2023--12}.
\newblock


\bibitem[Baskara(2023)]%
        {baskara2023navigating}
\bibfield{author}{\bibinfo{person}{FX~Risang Baskara}.} \bibinfo{year}{2023}\natexlab{}.
\newblock \showarticletitle{Navigating the Complexities and Potentials of Language Learning Machines in EFL Contexts: A Multidimensional Analysis}. In \bibinfo{booktitle}{\emph{ICON LATERALS 2023: Proceedings of the 4th International Conference Entitled Language, Literary, And Cultural Studies, ICON LATERALS 2023, 11-12 July 2023, Malang, Indonesia}}. European Alliance for Innovation, \bibinfo{pages}{39}.
\newblock


\bibitem[Berk(2009)]%
        {berk2009multimedia}
\bibfield{author}{\bibinfo{person}{Ronald~A Berk}.} \bibinfo{year}{2009}\natexlab{}.
\newblock \showarticletitle{Multimedia teaching with video clips: TV, movies, YouTube, and mtvU in the college classroom.}
\newblock \bibinfo{journal}{\emph{International Journal of Technology in Teaching \& Learning}} \bibinfo{volume}{5}, \bibinfo{number}{1} (\bibinfo{year}{2009}).
\newblock


\bibitem[Biri et~al\mbox{.}(2023)]%
        {biri2023assessing}
\bibfield{author}{\bibinfo{person}{Sairavi~Kiran Biri}, \bibinfo{person}{Subir Kumar}, \bibinfo{person}{Muralidhar Panigrahi}, \bibinfo{person}{Shaikat Mondal}, \bibinfo{person}{Joshil~Kumar Behera}, {and} \bibinfo{person}{Himel Mondal}.} \bibinfo{year}{2023}\natexlab{}.
\newblock \showarticletitle{Assessing the utilization of large language models in medical education: Insights from undergraduate medical students}.
\newblock \bibinfo{journal}{\emph{Cureus}} \bibinfo{volume}{15}, \bibinfo{number}{10} (\bibinfo{year}{2023}).
\newblock


\bibitem[Dimitrova et~al\mbox{.}(2017)]%
        {dimitrova2017using}
\bibfield{author}{\bibinfo{person}{Vania Dimitrova}, \bibinfo{person}{Antonija Mitrovic}, \bibinfo{person}{Alicja Piotrkowicz}, \bibinfo{person}{Lydia Lau}, {and} \bibinfo{person}{Amali Weerasinghe}.} \bibinfo{year}{2017}\natexlab{}.
\newblock \showarticletitle{Using learning analytics to devise interactive personalised nudges for active video watching}. In \bibinfo{booktitle}{\emph{Proceedings of the 25th conference on user modeling, adaptation and personalization}}. \bibinfo{pages}{22--31}.
\newblock


\bibitem[Fagbohun et~al\mbox{.}(2024)]%
        {fagbohun2024beyond}
\bibfield{author}{\bibinfo{person}{O Fagbohun}, \bibinfo{person}{NP Iduwe}, \bibinfo{person}{M Abdullahi}, \bibinfo{person}{A Ifaturoti}, {and} \bibinfo{person}{OM Nwanna}.} \bibinfo{year}{2024}\natexlab{}.
\newblock \showarticletitle{Beyond traditional assessment: Exploring the impact of large language models on grading practices}.
\newblock \bibinfo{journal}{\emph{Journal of Artifical Intelligence and Machine Learning \& Data Science}} \bibinfo{volume}{2}, \bibinfo{number}{1} (\bibinfo{year}{2024}), \bibinfo{pages}{1--8}.
\newblock


\bibitem[Ko et~al\mbox{.}(2023)]%
        {ko2023large}
\bibfield{author}{\bibinfo{person}{Dohwan Ko}, \bibinfo{person}{Ji~Soo Lee}, \bibinfo{person}{Wooyoung Kang}, \bibinfo{person}{Byungseok Roh}, {and} \bibinfo{person}{Hyunwoo~J Kim}.} \bibinfo{year}{2023}\natexlab{}.
\newblock \showarticletitle{Large language models are temporal and causal reasoners for video question answering}.
\newblock \bibinfo{journal}{\emph{arXiv preprint arXiv:2310.15747}} (\bibinfo{year}{2023}).
\newblock


\bibitem[Kosterelioglu(2016)]%
        {kosterelioglu2016student}
\bibfield{author}{\bibinfo{person}{Ilker Kosterelioglu}.} \bibinfo{year}{2016}\natexlab{}.
\newblock \showarticletitle{Student views on learning environments enriched by video clips.}
\newblock \bibinfo{journal}{\emph{Universal Journal of Educational Research}} \bibinfo{volume}{4}, \bibinfo{number}{2} (\bibinfo{year}{2016}), \bibinfo{pages}{359--369}.
\newblock


\bibitem[Lackmann et~al\mbox{.}(2021)]%
        {lackmann2021influence}
\bibfield{author}{\bibinfo{person}{Sergej Lackmann}, \bibinfo{person}{Pierre-Majorique L{\'e}ger}, \bibinfo{person}{Patrick Charland}, \bibinfo{person}{Caroline Aub{\'e}}, {and} \bibinfo{person}{Jean Talbot}.} \bibinfo{year}{2021}\natexlab{}.
\newblock \showarticletitle{The influence of video format on engagement and performance in online learning}.
\newblock \bibinfo{journal}{\emph{Brain Sciences}} \bibinfo{volume}{11}, \bibinfo{number}{2} (\bibinfo{year}{2021}), \bibinfo{pages}{128}.
\newblock


\bibitem[Lo and Hew(2019)]%
        {lo2019impact}
\bibfield{author}{\bibinfo{person}{Chung~Kwan Lo} {and} \bibinfo{person}{Khe~Foon Hew}.} \bibinfo{year}{2019}\natexlab{}.
\newblock \showarticletitle{The impact of flipped classrooms on student achievement in engineering education: A meta-analysis of 10 years of research}.
\newblock \bibinfo{journal}{\emph{Journal of Engineering Education}} \bibinfo{volume}{108}, \bibinfo{number}{4} (\bibinfo{year}{2019}), \bibinfo{pages}{523--546}.
\newblock


\bibitem[Mahmud et~al\mbox{.}(2011)]%
        {mahmud2011dissection}
\bibfield{author}{\bibinfo{person}{Waqas Mahmud}, \bibinfo{person}{Omar Hyder}, \bibinfo{person}{Jamaal Butt}, {and} \bibinfo{person}{Arsalan Aftab}.} \bibinfo{year}{2011}\natexlab{}.
\newblock \showarticletitle{Dissection videos do not improve anatomy examination scores}.
\newblock \bibinfo{journal}{\emph{Anatomical sciences education}} \bibinfo{volume}{4}, \bibinfo{number}{1} (\bibinfo{year}{2011}), \bibinfo{pages}{16--21}.
\newblock


\bibitem[Mannekote et~al\mbox{.}(2024)]%
        {mannekote2024large}
\bibfield{author}{\bibinfo{person}{Amogh Mannekote}, \bibinfo{person}{Adam Davies}, \bibinfo{person}{Juan~D Pinto}, \bibinfo{person}{Shan Zhang}, \bibinfo{person}{Daniel Olds}, \bibinfo{person}{Noah~L Schroeder}, \bibinfo{person}{Blair Lehman}, \bibinfo{person}{Diego Zapata-Rivera}, {and} \bibinfo{person}{ChengXiang Zhai}.} \bibinfo{year}{2024}\natexlab{}.
\newblock \showarticletitle{Large language models for whole-learner support: opportunities and challenges}.
\newblock \bibinfo{journal}{\emph{Frontiers in Artificial Intelligence}}  \bibinfo{volume}{7} (\bibinfo{year}{2024}), \bibinfo{pages}{1460364}.
\newblock


\bibitem[Mayer(2008)]%
        {mayer2008applying}
\bibfield{author}{\bibinfo{person}{Richard~E Mayer}.} \bibinfo{year}{2008}\natexlab{}.
\newblock \showarticletitle{Applying the science of learning: evidence-based principles for the design of multimedia instruction.}
\newblock \bibinfo{journal}{\emph{American psychologist}} \bibinfo{volume}{63}, \bibinfo{number}{8} (\bibinfo{year}{2008}), \bibinfo{pages}{760}.
\newblock


\bibitem[Mitrovic et~al\mbox{.}(2016)]%
        {mitrovic2016reflective}
\bibfield{author}{\bibinfo{person}{Antonija Mitrovic}, \bibinfo{person}{Vania Dimitrova}, \bibinfo{person}{Amali Weerasinghe}, {and} \bibinfo{person}{Lydia Lau}.} \bibinfo{year}{2016}\natexlab{}.
\newblock \showarticletitle{Reflective experiential learning: Using active video watching for soft skills training}. In \bibinfo{booktitle}{\emph{Proceedings of the 24th international conference on computers in education}}. Asia-Pacific Society for Computers in Education.
\newblock


\bibitem[Moreno and Mayer(2007)]%
        {moreno2007interactive}
\bibfield{author}{\bibinfo{person}{Roxana Moreno} {and} \bibinfo{person}{Richard Mayer}.} \bibinfo{year}{2007}\natexlab{}.
\newblock \showarticletitle{Interactive multimodal learning environments: Special issue on interactive learning environments: Contemporary issues and trends}.
\newblock \bibinfo{journal}{\emph{Educational psychology review}}  \bibinfo{volume}{19} (\bibinfo{year}{2007}), \bibinfo{pages}{309--326}.
\newblock


\bibitem[Moss(1983)]%
        {moss1983video}
\bibfield{author}{\bibinfo{person}{Robin Moss}.} \bibinfo{year}{1983}\natexlab{}.
\newblock \showarticletitle{Video: the educational challenge}.
\newblock  (\bibinfo{year}{1983}).
\newblock


\bibitem[Pan et~al\mbox{.}(2024)]%
        {pan2024integrating}
\bibfield{author}{\bibinfo{person}{Wuming Pan}, \bibinfo{person}{Ying Yang}, {and} \bibinfo{person}{Hao Yin}.} \bibinfo{year}{2024}\natexlab{}.
\newblock \showarticletitle{Integrating LLMs and software-defined resources for enhanced demonstrative cloud computing education in university curricula}.
\newblock \bibinfo{journal}{\emph{Journal of Infrastructure, Policy and Development}} \bibinfo{volume}{8}, \bibinfo{number}{12} (\bibinfo{year}{2024}), \bibinfo{pages}{8751}.
\newblock


\bibitem[Paramesti et~al\mbox{.}(2021)]%
        {paramesti2021effect}
\bibfield{author}{\bibinfo{person}{Aulika~Alya Paramesti}, \bibinfo{person}{Putri~Sekar Wiyati}, {and} \bibinfo{person}{Diah~Rahayu Wulandari}.} \bibinfo{year}{2021}\natexlab{}.
\newblock \showarticletitle{The Effect of Educational Video About Maternal Health Related to COVID-19 to Knowledge Level of Medical Student}.
\newblock \bibinfo{journal}{\emph{Jurnal Kedokteran Diponegoro (Diponegoro Medical Journal)}} \bibinfo{volume}{10}, \bibinfo{number}{2} (\bibinfo{year}{2021}), \bibinfo{pages}{124--127}.
\newblock


\bibitem[Pel{\'a}ez-S{\'a}nchez et~al\mbox{.}(2024)]%
        {pelaez2024impact}
\bibfield{author}{\bibinfo{person}{Iris~Cristina Pel{\'a}ez-S{\'a}nchez}, \bibinfo{person}{Davis Velarde-Camaqui}, {and} \bibinfo{person}{Leonardo~David Glasserman-Morales}.} \bibinfo{year}{2024}\natexlab{}.
\newblock \showarticletitle{The impact of large language models on higher education: exploring the connection between AI and Education 4.0}. In \bibinfo{booktitle}{\emph{Frontiers in Education}}, Vol.~\bibinfo{volume}{9}. Frontiers Media SA, \bibinfo{pages}{1392091}.
\newblock


\bibitem[Roy et~al\mbox{.}(2024)]%
        {roy2024assessing}
\bibfield{author}{\bibinfo{person}{Asitava~Deb Roy}, \bibinfo{person}{Ichchhit~Bharat Jaiswal}, \bibinfo{person}{Devendra~Nath Tiu}, \bibinfo{person}{Dipmala Das}, \bibinfo{person}{Shaikat Mondal}, \bibinfo{person}{Joshil~Kumar Behera}, {and} \bibinfo{person}{Himel Mondal}.} \bibinfo{year}{2024}\natexlab{}.
\newblock \showarticletitle{Assessing the Utilization of Large Language Model Chatbots for Educational Purposes by Medical Teachers: A Nationwide Survey From India}.
\newblock \bibinfo{journal}{\emph{Cureus}} \bibinfo{volume}{16}, \bibinfo{number}{11} (\bibinfo{year}{2024}), \bibinfo{pages}{e73484}.
\newblock


\bibitem[Tarchi et~al\mbox{.}(2021)]%
        {tarchi2021learning}
\bibfield{author}{\bibinfo{person}{Christian Tarchi}, \bibinfo{person}{Sonia Zaccoletti}, {and} \bibinfo{person}{Lucia Mason}.} \bibinfo{year}{2021}\natexlab{}.
\newblock \showarticletitle{Learning from text, video, or subtitles: A comparative analysis}.
\newblock \bibinfo{journal}{\emph{Computers \& Education}}  \bibinfo{volume}{160} (\bibinfo{year}{2021}), \bibinfo{pages}{104034}.
\newblock


\bibitem[Teoh et~al\mbox{.}(2023)]%
        {teoh2023machine}
\bibfield{author}{\bibinfo{person}{Chin-Wei Teoh}, \bibinfo{person}{Sin-Ban Ho}, \bibinfo{person}{Khairi~Shazwan Dollmat}, {and} \bibinfo{person}{Chuie-Hong Tan}.} \bibinfo{year}{2023}\natexlab{}.
\newblock \showarticletitle{Machine Learning Prediction Model for Early Student Academic Performance Evaluation in Video-Based Learning}.
\newblock \bibinfo{journal}{\emph{International Journal}} \bibinfo{volume}{10}, \bibinfo{number}{2} (\bibinfo{year}{2023}), \bibinfo{pages}{1529--1544}.
\newblock


\bibitem[Zamanifard et~al\mbox{.}(2025)]%
        {zamanifard2025undergraduate}
\bibfield{author}{\bibinfo{person}{Samaneh Zamanifard}, \bibinfo{person}{Sajad Goudarzi}, \bibinfo{person}{Masoumeh Soleimani}, {and} \bibinfo{person}{Andrew Robb}.} \bibinfo{year}{2025}\natexlab{}.
\newblock \showarticletitle{Undergraduate Students’ Perceptions of Large Language Models in Higher Education: Self-Perceived Benefits, Reliance, and Accuracy Assessment}.
\newblock  (\bibinfo{year}{2025}).
\newblock


\bibitem[Zhang et~al\mbox{.}(2023)]%
        {zhang2023video}
\bibfield{author}{\bibinfo{person}{Hang Zhang}, \bibinfo{person}{Xin Li}, {and} \bibinfo{person}{Lidong Bing}.} \bibinfo{year}{2023}\natexlab{}.
\newblock \showarticletitle{Video-llama: An instruction-tuned audio-visual language model for video understanding}.
\newblock \bibinfo{journal}{\emph{arXiv preprint arXiv:2306.02858}} (\bibinfo{year}{2023}).
\newblock


\bibitem[Zhou et~al\mbox{.}({[n.\,d.]})]%
        {zhou2404survey}
\bibfield{author}{\bibinfo{person}{Pengyuan Zhou}, \bibinfo{person}{Lin Wang}, \bibinfo{person}{Zhi Liu}, \bibinfo{person}{Yanbin Hao}, \bibinfo{person}{Pan Hui}, \bibinfo{person}{Sasu Tarkoma}, {and} \bibinfo{person}{Jussi Kangasharju}.} \bibinfo{year}{[n.\,d.]}\natexlab{}.
\newblock \showarticletitle{A survey on generative AI and LLM for video generation, understanding, and streaming 2024}.
\newblock \bibinfo{journal}{\emph{URL https://arxiv. org/abs/2404.16038}} (\bibinfo{year}{[n.\,d.]}).
\newblock


\bibitem[Zhui et~al\mbox{.}(2024)]%
        {zhui2024impact}
\bibfield{author}{\bibinfo{person}{Li Zhui}, \bibinfo{person}{Nina Yhap}, \bibinfo{person}{Liu Liping}, \bibinfo{person}{Wang Zhengjie}, \bibinfo{person}{Xiong Zhonghao}, \bibinfo{person}{Yuan Xiaoshu}, \bibinfo{person}{Cui Hong}, \bibinfo{person}{Liu Xuexiu}, \bibinfo{person}{Ren Wei}, {et~al\mbox{.}}} \bibinfo{year}{2024}\natexlab{}.
\newblock \showarticletitle{Impact of large language models on medical education and teaching adaptations}.
\newblock \bibinfo{journal}{\emph{JMIR Medical Informatics}} \bibinfo{volume}{12}, \bibinfo{number}{1} (\bibinfo{year}{2024}), \bibinfo{pages}{e55933}.
\newblock


\end{thebibliography}


\end{document}